\documentclass[11pt,a4paper]{article}
\usepackage[hyperref]{emnlp-ijcnlp-2019}
\usepackage{times}
\usepackage{latexsym}
\usepackage{graphicx}
\usepackage{url}
\usepackage{multirow}
\usepackage{hyperref}
\usepackage{float}
\usepackage{booktabs}
\usepackage{footnote}
\usepackage{threeparttable}
\usepackage{longtable}
\usepackage[justification=centering]{caption}
\usepackage{xcolor}
\usepackage{makecell}
\makesavenoteenv{tabular}
\makesavenoteenv{table}
\usepackage{todonotes}
\usepackage[shortlabels]{enumitem}
\setlist[enumerate]{leftmargin=*}
\setlist{nolistsep}

\usepackage{amsmath}
\usepackage{amssymb}
\usepackage{amsthm}
\usepackage{adjustbox}

\newtheorem{definition}{Definition}
\newtheorem{conjecture}{Conjecture}

\title{How Sequence-to-Sequence Models Perceive Language Styles?}

\author{Ruozi Huang \\
  {\tt 17212010012@fudan.edu} \\
  Mi Zhang \\ 
  {\tt mi\_zhang@fudan.edu.cn} \\\And
  Xudong Pan \\
  {\tt 18110240010@fudan.edu.cn} \\
  Beina Sheng\\
  {\tt 16307130370@fudan.edu.cn}}

\begin{document}
\maketitle
\begin{abstract}
Style is ubiquitous in our daily language uses, while \emph{what is language style to learning machines}? In this paper, by exploiting the second-order statistics of semantic vectors of different corpora, we present a novel perspective on this question via \textit{style matrix}, i.e. the covariance matrix of semantic vectors, and explain for the first time how Sequence-to-Sequence models encode style information innately in its semantic vectors. As an application, we devise a learning-free text style transfer algorithm, which explicitly constructs a pair of transfer operators from the style matrices for style transfer. Moreover, our algorithm is also observed to be flexible enough to transfer out-of-domain sentences. Extensive experimental evidence justifies the informativeness of style matrix and the competitive performance of our proposed style transfer algorithm with the state-of-the-art methods.
\end{abstract}

\section{Introduction} \label{sec:intro}
Different corpus may present different language styles, featured by variations in attitude, tense, word choice, \textit{et cetera}. As human beings, we always have an intuitive perception of style differences in texts. In the literature of linguistics, there also developed a number of mature theories for characterizing style phenomena in our daily lives \citep{bell1984language, coupland2007style, ray2014style}. However, with decades of advancement of machine learning techniques in Natural Language Processing (NLP), an interesting and fundamental question still remains open: \textit{How is style information encoded by learning models?}

In this paper, we share our novel observations for this question in a specified version, that is, in what approach Sequence-to-Sequence (\textit{seq2seq}; \citealt{sutskever2014sequence}), a prestigious neural network architecture widely used in NLP and representation learning \citep{bahdanau2014neural, li2015hierarchical,kiros2015skip}, encodes language styles. 

\begin{figure}
\centering
\includegraphics[width=0.45\textwidth]{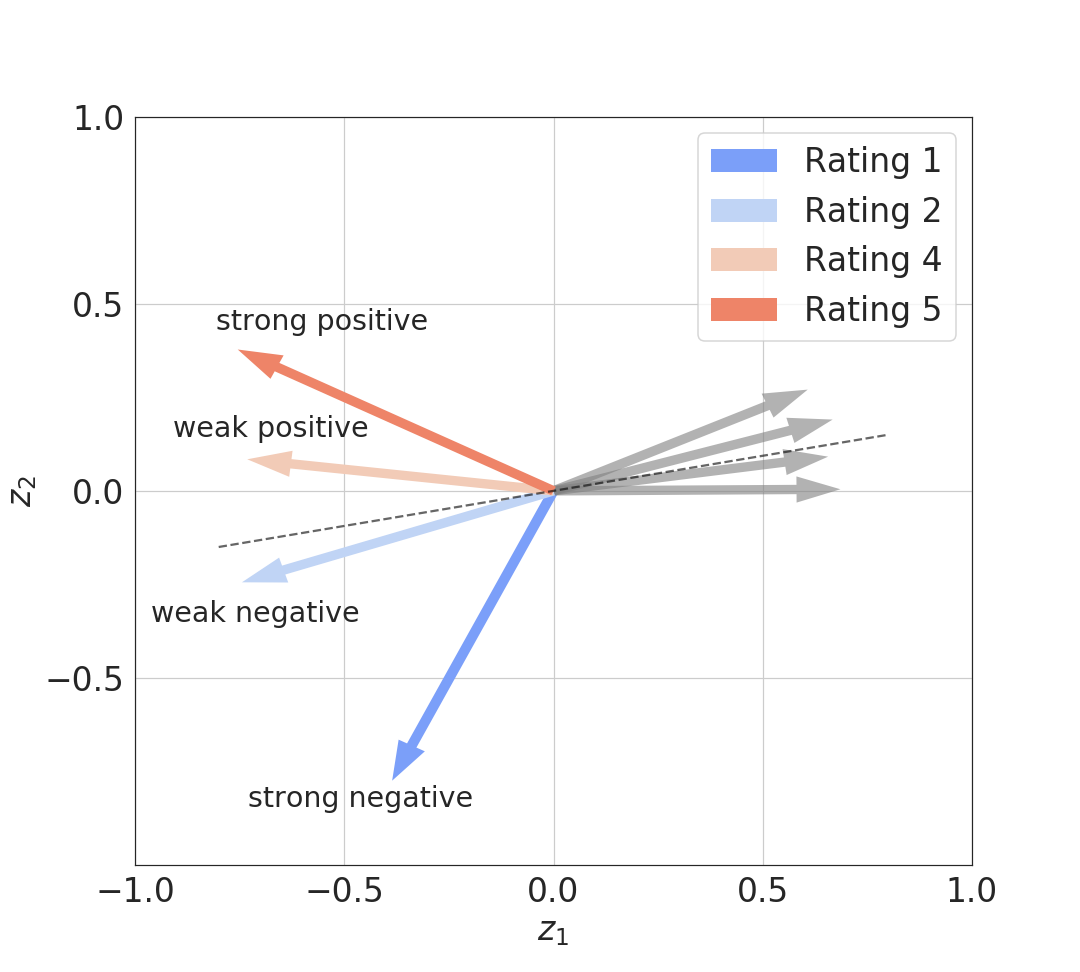}
\caption{Visualization of the first two eigenvectors of style matrices on four subcorpora with ratings ranging in $1$, $2$, $4$, $5$, collected from Yelp review dataset. The reviews with higher rating are more positive in attitude and vice versa.}
\label{fig:eig}
\end{figure}

In our preliminary studies, we applied a typical seq2seq model as an autoencoder to learn semantic vectors for sentences from Yelp review dataset \footnote{https://www.yelp.com/dataset/challenge} and we strikingly made the following observation: After calculating respectively the covariance matrices of the semantic vectors of reviews with different attitude polarity and intensity, we found their \textit{second eigenvectors}, i.e. eigenvectors with the second largest eigenvalues, were roughly grouped in two parts according to their polarity and meanwhile showed slight difference according to their intensity, as in the colored part of Fig. \ref{fig:eig}, while the \emph{first eigenvectors} illustrated in gray formed a single cluster, as if they captured certain common attributes of the Yelp corpus, e.g. casual word choices. This phenomenon suggests the covariance matrices have probably encoded the language style in an informative way. Based on this observation, we provide the notion of style matrix and investigate a number of far-reaching implications brought by style matrix in the remainder of this paper.

\begin{conjecture}[Style Matrix]\label{conj:style_mat}
The style of a corpus is encoded in the covariance matrix of its semantic vectors, which is called \emph{style matrix}.
\end{conjecture}

For the best of our knowledge, our research question and conjecture are quite novel and there barely exist any relevant works before. The most related works are probably the recent studies on text style transfer, a task which was first investigated by \citet{shen2017style} for converting a given corpus in one style to another. Although most of them have not discussed style at a fundamental level, we mainly identify three related perspectives in existing style transfer methods.

\begin{enumerate}[A.]
\item \emph{Discrete Label}: A major proportion of the state-of-the-art style transfer methods simply view the corpus style as discrete labels. For instance, the style of positive and negative reviews in Yelp dataset are respectively assigned with binary labels \citep{shen2017style,hu2017toward,chen2018adversarial}. 

\item \emph{Style Embedding}: Instead of discrete labels, some methods propose to learn \emph{semantic-independent} embedding vectors as distributed representations of style \citep{fu2018style}.

\item \emph{Lexicon-Based}: Other works suggest use the most significant lexical units, i.e. those serving as major factors to style classification decision, as representatives of text style \citep{li2018delete,xu2018unpaired}.
\end{enumerate}


In principle, our main conjecture is better compatible with linguistic aspects of style phenomenon than the aforementioned perspectives, especially in the following aspects.

\begin{enumerate}
\item \textit{Style is a statistical phenomenon.} According to the variationist's view in sociolinguistics \cite{coupland2007style}, style emerges from variation of language usages and is always a global phenomenon rather than a property of single sentence. In our conjecture, style matrix by definition reflects the covariance of the corpus, while Perspective C could only characterize the sentence-level style.
\item \textit{Style is inherent in semantics.} As \citet{ray2014style} suggests, expression often helps to form meaning. In our conjecture, style matrix is an explicit function of semantic vectors, while Perspective B improperly assumed style embedding's independence on semantic.
\item \textit{Style is multi-modal.} Usually, we recognize style in texts from many different aspects \cite{bell1984language}. For instance, \textit{I am very unhappy with this place} is negative in attitude and meanwhile present in tense. Moreover, one can also recognize slight differences in style intensity for both negative sentences, i.e. with/without \textit{very} in the example above. With experiments, we show style matrix is able to distinguish various intensity level of style (\S \ref{sec:style_intensity}) and capture multiple styles in one corpus (\S \ref{sec:multi_style}), while Perspective A is impotent to characterize these delicate style differences by discrete labels.
\end{enumerate}

In practice, based on the notion of style matrix, we propose a novel algorithm called \emph{Neutralization-Stylization} (NS) for unpaired text style transfer. Given style matrices of source and target corpora obtained from a pre-trained seq2seq autoencoder, our algorithm works in a fully learning-free manner by first preparing a pair of matrix transform operators from the style matrices. After the preparation, it simply applies these operators to the style matrix of given corpus to accomplish text style transfer on the fly. By introducing additional style information as supervision on the learning process of the seq2seq autoencoder, we observe NS algorithm can achieve comparable performance with the state-of-the-art style transfer methods on each standard metric. Moreover, the flexibility of our method is further demonstrated by its ability to control the style of unlabeled sentences from other domains, i.e. \emph{out-of-domain text style transfer}, which we propose as a much challenging task to foster future researches.

In summary, our contributions are as follows:
\begin{itemize}[leftmargin=*]
    \item We present the notion of style matrix as an informative delegate to language style and explains for the first time how seq2seq models encode language styles (\S \ref{sec:style_mat}).
    \item With the aid of style matrix, we devise Neutralization-Stylization as a learning-free algorithm for text style transfer among binary, multiple and mixed styles (\S \ref{sec:transfer}), which achieves competitive performance compared with the state-of-the-art methods (\S \ref{sec:comparison}).
    \item We introduce the challenging out-of-domain text style transfer task to further prove the flexibility of our proposed method (\S \ref{sec:ood}).
\end{itemize}

\section{Style Matrix} \label{sec:style_mat}
\subsection{A General Framework for Style Matrix Extraction}
Given a corpus $X=\{x^{(1)},x^{(2)},\hdots,x^{(N)}\}$, where $x^{(i)}$ is a sequence of tokens, we wonder whether there exists an explicit way to extract the global style of $X$ with no other external knowledge. Inspired from the variationist's approach to language style in the context of sociolinguistic \citep{coupland2007style}, we suggest exploiting the second-order statistics of semantics, specifically the covariance matrix, as an informative representation of the corpus style (Conjecture \ref{conj:style_mat}). Somewhat coincidentally, a similar viewpoint on visual style has been investigated in the computer vision community recently \citep{gatys2015texture}.

Due to the discrete essence of language, to compute the covariance matrix is not directly applicable to raw representations such as those in one-hot scheme. In order to fulfill the statement in our main conjecture, we require the semantic vector to be both \emph{distributed} and nearly \emph{lossless}. The former property requires the semantic of the original sentence can be compressed into a latent vector, while the latter requires the original sentence can be near-optimally reconstructed from the semantic vector alone.

Formally, we first convert $X$ into distributed representations with a mapping $E$ (i.e. \emph{encoder}) from $X$ to $\mathbb{R}^{d}$, a $d$-dimensional real-valued vector space. In order to guarantee $E$ is lossless, we further require the existence of a reverse mapping $D$ (i.e. \emph{decoder}) from $\mathbb{R}^{d}\to{X}$ which satisfies $D\circ{E} = I_{X}$, the identity mapping on the corpus. Once these conditions satisfied, we call the distributed representations $Z$, which consists of $\{z^{(i)}\}_{i=1}^{N}$ s.t. $z^{(i)} = E(x^{(i)})$, the \emph{semantic vectors} of corpus $X$. As a slight abuse of notation, we also use $Z$ to represent the semantic vecotrs in matrix form, i.e. $Z \doteq [z^{(1)}, \hdots, z^{(N)}]\in\mathbb{R}^{d\times{N}}$. Based on these notations, we provide the formal counterpart to Conjecture \ref{conj:style_mat} as follows.

\begin{definition}[Style Matrix]\label{def:style_mat}
Given corpus $X$ with a semantic encoder $E$ satisfying the requirements above, we define the style matrix $S_{X}\in\mathbb{R}^{d\times{d}}$ as 
\begin{equation}\label{eq:style_mat}
S_X = \frac{1}{N-1}\hat{Z}_X\hat{Z}_X^{T}
\end{equation}
where $\hat{Z}_X$ denotes $Z_X$ after being centered, i.e. $\hat{Z}_X \doteq Z_X - \bar{z}_X\mathbf{1}_{d}^{T}$ and $\bar{z}_X \doteq\frac{1}{N}\sum_{i=1}^{N}z_X^{(i)}$.
\end{definition}

In recent studies of sentence embedding (e.g. \citealt{le2014distributed, conneau2017supervised, pagliardini2017unsupervised}), there indeed exist various existing choices for implementing encoder $E$. However, as most of them do not have an explicit notion of the decoder, the extracted style matrix would therefore not be able to be further utilized in downstream style transfer tasks. Therefore, in the next section, we propose to leverage the power of seq2seq paradigm \citep{sutskever2014sequence} as a practical tool for extracting highly informative style matrix and meanwhile, facilitates style transfer tasks with the simultaneously trained decoder module. A detailed implementation is provided below.

\subsection{Case Study: Seq2seq for Style Matrix Extraction}\label{sec:style_mat_extraction}
As an overview, we implement the encoder $E$ in Definition \ref{def:style_mat} with the encoder module of a seq2seq model, while its decoder module $G$ learns alongside $E$ under the reconstruction loss to guarantee the original semantic is largely preserved in the obtained semantic vectors $E(X)$. Given a sentence $\mathbf{x} = \{w_1, \hdots, w_T\}$ with each token $w_i$ from a vocabulary $\mathcal{V}$, we propose the learning process for style matrix extraction below.

For the encoder module, we use a recurrent neural network with Gated Recurrent Unit (GRU; \citet{cho2014learning}), i.e. $\text{GRU}_{E}$, to encode x into hidden state vectors with
\begin{equation}
\mathbf{h}=\text{GRU}_{E}(\mathbf{x})
\end{equation}
where $\mathbf{h}=\{h_1,h_2,\ldots,h_T\}$ contains all the hidden states calculated by the GRU encoder.  

Next, viewing the last hidden state $h_T$ of the encoder as the semantic vector $z$, we further require the decoder can reconstruct $x$ based on $z$ token by token with a GRU (denoted as $\text{GRU}_{D}$). Formally, at each step $t$, $\text{GRU}_{D}$ takes the generated token $y_{t}$ and the previous hidden state $s_{t-1}$ as input to calculate the current state by
\begin{equation}
s_t=\text{GRU}_{D}(y_t,s_{t-1})
\end{equation}
where the initial state is set as $z$.

Subsequently, with a linear projection layer followed by a softmax transformation, the distribution of the next token $y_{t+1}$ over the vocabulary is calculated as
\begin{equation}
P(y_{t}|\mathbf{x}, y_{<t}) = \text{softmax}(Ws_t)
\end{equation}
where $W$ is a learnable matrix in $\mathbb{R}^{|\mathcal{V}|\times{d}}$. 

By convention of unsupervised learning \citep{lecun2015deep}, we set the reconstruction objective as the categorical cross entropy between the input sequence $\mathbf{x}$ and the distribution of the reconstructed sequence $\mathbf{y}$. In practice, we further apply the scheduled sampling technique to accelerate the aforementioned learning process \citep{bengio2015scheduled}. 

It is worth to notice, in our implementation of seq2seq for autoencoding, we have intentionally avoided the usage of attention mechanism \citep{luong2015effective}. It is mainly because, with the attention mechanism, information flow from encoder to decoder is not limited to the semantic vector $z$. For example, the reconstruction process is otherwise also dependent on the context vector. Therefore, although attention mechanism can bring optimal reconstruction loss even with small hidden state size, it may cause potential semantic loss and therefore compromise the quality of the extracted style matrix. 

As a final remark, we demonstrate our method above with GRU modules only for the sake of concreteness. Besides GRU, there are various available recurrent architectures for implementing $E, G$, such as vanilla recurrent unit \citep{rumelhart1985learning}, Long Short-Term Memory network (LSTM; \citet{hochreiter1997long}) and their bidirectional or stacked variants \citep{jurafsky2000speech}. In experiments, we also report results with several typical architectures as a comprehensive self-comparison.

\section{Style Transfer with Style Matrix} \label{sec:transfer}
In this section, we propose a novel algorithm called \emph{Neuralization-Stylization} (NS) for unpaired text style transfer by directly \emph{aligning} the style matrix of one corpus to the other with a pair of plug-and-play matrix operations. To achieve competitive performance as the state-of-the-art style transfer methods, we further augment the unsupervised style matrix extraction process in Section \ref{sec:style_mat_extraction} by introducing human-defined style information as external supervision.

\subsection{Neutralization-Stylization algorithm}

As a covariance matrix in essence, style matrix $S_X$ can be factorized into the following form due to its positive semi-definiteness 
\begin{equation}\label{eq:eigen}
S_X=P_X \Lambda_X P_X^{T} 
\end{equation}
where $\Lambda_X$ is a diagonal matrix consisting of its eigenvalues and $P_X$ is an orthogonal matrix formed by its eigenvectors \citep{meyer2000matrix} .

Given two corpora $X$ and $Y$ and a seq2seq autoencoder $(E, G)$ pretrained on $X \cup Y$ as a larger corpus, we calculate Eq. \ref{eq:style_mat} respectively on $X, Y$ to obtain the style matrices $S_X, S_Y$. Using eigenvalue decomposition in Eq. \ref{eq:eigen}, we next introduce a pair of Neutralization and Stylization operators, which can be easily used for on-the-fly text style transfer in a plug-and-play manner. Note both operators are defined on a set of semantic vectors rather than a single embedding, which highly corresponds to the statistical essence of language style \citep{coupland2007style}. 

\subsubsection{Plug-and-Play Style Transfer Operators}

\textbf{Neutralization.} Neutralization operator is used to remove the style characteristic of corpus $X$ from a set of semantic vectors $Z$. Formally, in the spirit of Zero-phase Component Analysis (ZCA) \citep{bell1997edges}, neutralization operator is defined as
\begin{equation}
N_XZ =P_X\Lambda_X^{-\frac{1}{2}}P_X^{\mathrm{T}}(Z - \bar{z}_X\mathbf{1}_{d}^{T})
\end{equation}

An intuitive way to understand how it works is by replacing $Z$ directly with the semantic vectors of $X$. It is easy to check: $N_XZ_X$ has its style matrix as $\mathbf{I}_{d\times{d}}$, which means the dimensions of semantics become uncorrelated after neutralization.
 
\noindent\textbf{Stylization.} Stylization transformation is used to add the style characteristic of corpus $Y$ to 
a set of semantic vectors $Z$ by reestabilishing the correlation among dimensions of semantics, which, with inspirations from \citet{hossain2016whitening}, is defined as
\begin{equation}
S_YZ=P_Y\Lambda_Y^{\frac{1}{2}}P_Y^{\mathrm{T}}Z +  \bar{z}_Y\mathbf{1}_{d}^{T}
\end{equation}

Similarly, by stylizing a neutral set of semantic vectors $Z$ (i.e. $ZZ^{T} = \mathbf{I}_{d\times{d}}$), we can easily check $S_YZ$ has the same style matrix as that of corpus $Y$, which hence demonstrates the properness of $S_Y$. 
\subsubsection{On-the-Fly Text Style Transfer}
With the well-defined neutralization and stylization operators, our proposed learning-free NS algorithm works straightforwardly by: (1) encoding $x$ with $E$; (2) applying prepared $(N, S)$ operators successively; (3) decoding the semantic vector with $D$. Formally, the target sentence $y$ is calculated as 
\begin{equation}
    y = D(S_YN_XE(x))
\end{equation}

Moreover, thanks to the flexibility of style matrix perspective and NS algorithm, we can even conduct \emph{out-of-domain style transfer}, where the input sentence $x$ not necessarily comes from corpus $X$ or has style labels. For details, we present an interesting case study on out-of-domain style transfer between Yelp and Amazon datasets in Section \ref{sec:ood}.

\subsection{Incorporate Human-Defined Style Label} \label{sec:style_label}
In practice, we notice the performance of NS algorithm with raw style matrix is not competitive with the state-of-the-art methods specified on this task. We speculate the main reason lies in: Style matrix is highly informative and probably incorporates even the most delicate aspect of style of the underlying corpus. Therefore, its unsatisfactory performance on style transfer task implies the corpus actually has other latent attributes of style besides the human-defined ones, as we have illustrated with the Yelp example in Section \ref{sec:intro} by its clustered first eigenvectors (gray arrows in Fig. \ref{fig:eig}). 

To enhance the quality of style transfer, we suggest to augment the style matrix extraction process with human-defined attribute (e.g. attitude). Concretely, we propose to train the encoder $E$ of the seq2seq model in a semi-supervised way by adding a nonlinear binary classifier $C: \mathbb{R}^{d}\to\{0, 1\}$ on the semantic space, which provides supervision signal simultaneously with the original unsupervised reconstruction process. Formally, given semantic vector $z$, we define the classifier as
\begin{equation}
C(z)=\sigma(w^{T}\text{vec}(zz^T))
\end{equation}
where $w\in R^{d^2}$ is the trainable parameter and $\sigma$ is the sigmoid activation.

Noticeably, under both scenarios, our text style transfer algorithm is learning-free because: we only need to pretrain a seq2seq model, either in fully unsupervised or semi-supervised way, to obtain a pair of encoder and decoder and prepare the $(N, S)$ operators with several matrix operations. Without time-consuming adversarial training (e.g. \citet{shen2017style}), our augmented method achieved competitive transfer performance on each standard metric (\S \ref{sec:comparison}).

\section{Experiments\footnote{Code is provided at \url{https://bit.ly/2QgEUNE}}}
\subsection{Overall Settings}
\noindent\textbf{Datasets.}  We used the following two standard benchmark datasets for empirical studies.

\textbf{Yelp:} The Yelp dataset collected the reviews to restaurants on Yelp. Each sentence is associated with an integer rating ranging from $1$ to $5$, where a higher score implies the more positive of the corresponding review's attitude and vice versa. We treated the attitude of reviews with ratings above $3$ as positive while those below $3$ as negative. 

\textbf{Amazon:} The Amazon dataset contains the product reviews on Amazon. Each sentence is originally labeled with positive or negative attitudes \citep{he2016ups}.

With an automatic tense analysis tool \citep{ramm2017annotating}, we annotated the tense attribute for sentences in Yelp and Amazon as an additional style factor. We filtered out the sentences which were not in past and present tense and split each processed dataset into train, validation and test sets. For statistics, please refer to Appendix A.

\noindent\textbf{Evaluation Metric.} 
We evaluated the performance of style transfer on the following two standard metrics.

\textbf{Accuracy (Acc.)}: In order to evaluate whether the transferred sentences have the desired style, we followed the evaluation method in \citet{shen2017style} by pretraining a style classifier on the training set and utilizing its classification accuracy on the transferred sentences as a metric. Specifically, we used the TextCNN model \citep{kim2014convolutional} as a style classifier. 

\textbf{BLEU}: In order to evaluate the quality of content preservation, we used the BLEU score \cite{papineni2002bleu} between the generated and the source sentences as a measure. Intuitively, a higher BLEU score primarily indicates the model has a stronger ability to preserve content by copying style-neutral words from the source sentence. 

To evaluate the overall performance of style transfer quality, we also calculated the geometric mean (i.e. \textit{G-Score}) and arithmetic mean (i.e. \textit{Mean}) of Acc. and BLEU metrics.

\begin{table}
\caption{Results of style transfer with NS algorithm on corpora pairs with different style contrast levels.}\label{tab:intensity_level}
\centering
\begin{adjustbox}{width=0.45\textwidth}
\begin{tabular}{c|c|c}
\hline
NS Operators From &Acc. (Baseline\footnote{The error rate of original sentences in validation set is provided as baseline, so as in Table \ref{tab:multiclass}})&BLEU\cr
\hline
(R1, R5)&39.48 (2.67)&36.97\cr
(R1 $\cup$ R2, R4 $\cup$ R5) &35.84 (2.67)&39.44\cr
(R2, R4)&32.82 (2.67)&41.08\cr
\hline
\end{tabular}
\end{adjustbox}
\label{tab_star}
\end{table}

\noindent\textbf{Implementation Details.}  We embedded words into distributed representations (with dimension $d = 300$) using CBOW \cite{mikolov2013distributed} and froze the word embeddings during the training process. We implemented the seq2seq model with (1) GRU of $300$ hidden units, (2) LSTM of $150$ hidden units and (3) bi-directional GRU of both $150$ forward and backward hidden units. For (2), we concatenated the final hidden state and cell state to form the $300$-dimensional semantic vector, while for (3), we concatenated the forward and backward final hidden states. We trained each seq2seq model on the training set with Adam optimizer \cite{kingma2014adam} and performed style transfer on the validation set. We set the weight of reconstruction loss and classification loss as 10:1. As observed in Section \ref{sec:comparison}, the informativeness of style matrix was insensitive to different choices of recurrent architectures and hence we only report the results of GRU implementation in other parts.

\subsection{Explore the Styles of Yelp}\label{sec:4_3}
As is discussed in Section \ref{sec:intro}, the notion of style matrix conforms to the linguistic aspects that style is innate in semantics and is multi-modal. To demonstrate style matrix can indeed capture these delicate style phenomena, we first mixed up all the reviews on Yelp with different ratings and trained a seq2seq model with reconstruction loss only. We then divided the corpus into several sub-corpora with well-designed criteria. Finally, we performed text style transfer with $(N,S)$ operators prepared respectively with these pairs of sub-corpora. Detailed results and analyses are followed in each part.
\begin{table*}[ht]
\centering
\captionsetup{justification=centering}
\caption{Performance of different style transfer methods on standard benchmarks.}\label{tab:performance}
 \begin{adjustbox}{width=0.8\textwidth}
\begin{tabular}{c|cccc|cccc}
\toprule
\multirow{2}{*}{Model}&
\multicolumn{4}{c}{ Yelp}&\multicolumn{4}{c}{ Amazon}\cr
\cmidrule(lr){2-5} \cmidrule(lr){6-9}
&Acc.&BLEU&G-Score&Mean&Acc.&BLEU&G-Score&Mean\cr
\midrule
Test Set&97.48&-&-&-&80.97&-&-&-\cr
Cross-Aligned&\textbf{83.78}&12.69&32.37&46.73&60.84&8.56&22.82&34.70\cr
Style-Embedding&6.34&\textbf{85.14}&23.23&45.74&29.21&\textbf{68.14}&\textbf{44.61}&48.68\cr
NS-GRU&80.33&13.43&\textbf{32.85}&\textbf{46.88}&\textbf{79.50}&12.97&32.11&46.23\cr
NS-LSTM&78.07&12.38&31.10&45.23&74.30&15.90&34.37&45.10\cr
NS-BiGRU&72.02&12.60&30.12&42.31&74.04&25.23&43.22&\textbf{49.64}\cr
\bottomrule
\end{tabular}
\end{adjustbox}
\end{table*}

\subsubsection{Style Intensity} \label{sec:style_intensity}
\begin{figure}
\centering{
\includegraphics[width=0.5\textwidth]{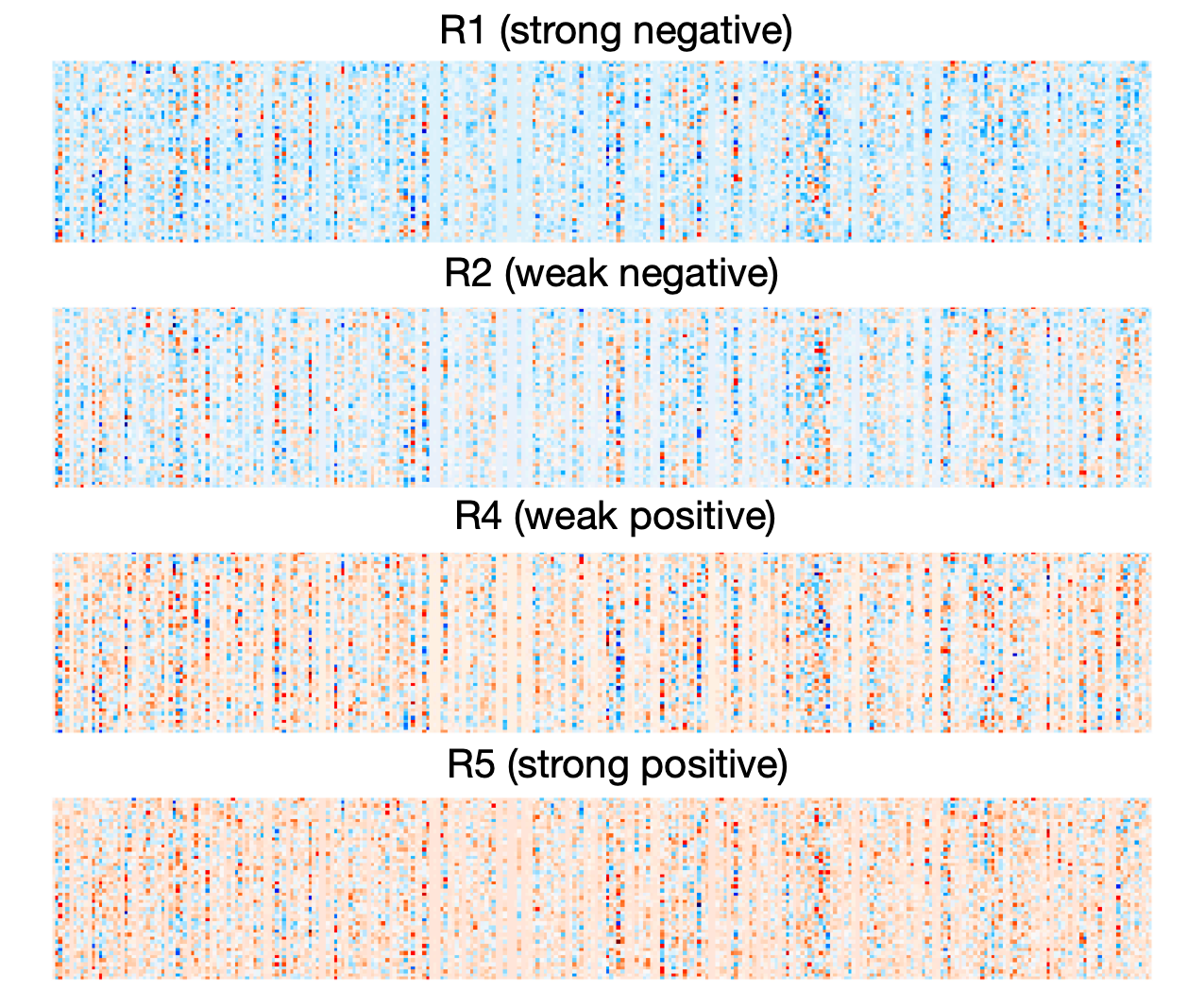}
}
\caption{Heatmap of the first $50$ eigenvectors of style matrices on Yelp subcorpora with different intensity levels of attitude, from R1 the most negative corpus to R5 the most positive one (better viewed in color).}
\label{fig:heatmap}
\end{figure}

We collected four corpora which contained sentences respectively with rating $1$, $2$, $4$ and $5$ (denoted as R1, R2, R4, R5) and discarded the neutral sentences with rating $3$. The former two sub-corpora have the same polarity of attitude (i.e. negative) but with different intensity and so as the latter two. For visualization, Fig. \ref{fig:heatmap} plots the first $50$ eigenvectors of each style matrix, which shows a recognizable color gradience from the most negative corpus (with rating $1$) to the most positive corpus (with rating $5$).

Subsequently, we constructed three sets of $(N, S)$ operators respectively from stylistic pairs (R1, R5), (R1 $\cup$ R2, R4 $\cup$ R5) and (R2, R4), in the decreasing order of style contrast level. We performed style transfer on the same validation set with the three sets of prepared $(N, S)$ operators. The results are reported in Table \ref{tab:intensity_level}. As we can see, the style transfer quality of each set of $(N, S)$ operators were positively related to the degree of style contrast and we suggest this phenomenon as an implicit validation for the informativeness of style matrix on capturing slight difference in style intensity.

\subsubsection{Multiple Styles} \label{sec:multi_style}
\begin{table}[ht]
\caption{Results of style transfer with NS algorithm on corpora pairs with multiple styles.} \label{tab:multiclass}
\centering
\begin{adjustbox}{width=0.35\textwidth}
\begin{tabular}{c|c|c}
\hline
 &Acc. (Baseline)&BLEU\cr
\hline
Attitude &29.56 (2.67)&62.04\cr
Tense &65.94 (2.73)&52.28\cr
\hline
\end{tabular}
\end{adjustbox}
\end{table}

Based on the attitude and tense annotations on Yelp, we partitioned the original corpus into two pairs of sub-corpora, namely the attitude pair \textit{(positive, negative)} and the tense pair \textit{(present, past)}. Correspondingly, we calculated attitude (tense) transform operators respectively on each pair and applied the prepared operators to transfer the target attribute with the other style attribute fixed. We report the transfer performance of NS algorithm in Table \ref{tab:multiclass}, which empirically proved style matrix can simultaneously capture multiple style attributes.

\subsection{Unpaired Text Style Transfer} \label{sec:comparison}

In this part, we compared the performance of NS algorithm with the state-of-the-art methods on Yelp and Amazon datasets. We trained a seq2seq model in the semi-supervised way as described in Section \ref{sec:style_label} and transferred attitude of sentences with NS algorithm. We chose the following representative state-of-the-art style transfer methods as baselines.

\textbf{Cross-Aligned:} This method assumes a shared latent content distribution across the corpora with different styles and leverages refined alignment of latent representations to perform style transfer \citep{shen2017style} . 

\textbf{Style-Embedding:} This method learns separate content representations and style representations using adversarial networks. With the style information embed into distributed vector representations, one single decoder is trained for different corpora \citep{fu2018style}.  

\begin{figure}[ht]
\centering{
\includegraphics[width=0.4\textwidth]{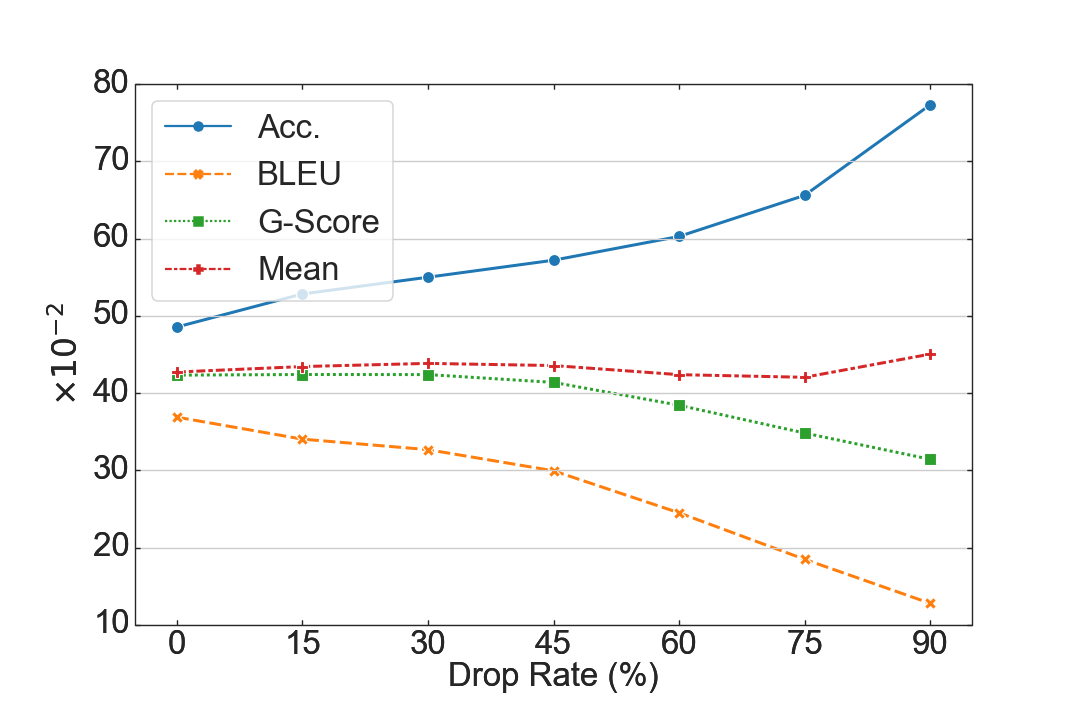}
}
\caption{Performance of NS algorithm with different sentences' drop rates for preparing transfer operators.}
\label{fig:line}
\end{figure}

As observed in Sec. \ref{sec:style_intensity}, the transform operators have a stronger transfer capability when generated from a pair of corpora with higher style contrast, which inspires us to further enhance the performance of NS algorithm by removing sentences with low confidence judged by the simultaneously trained style classifier. Fig. \ref{fig:line} plots the model performance on different metrics over drop rates ranging from $0$ to $0.9$ with a fixed stride $0.15$. 

As we can see, the increase in drop rate caused an increase of Acc. and decrease of BLEU score. We speculate it is inevitable due to the tight interdependence between style and semantics. The result at $90\%$ drop rate provides further evidence on this phenomenon, that is, to change the style of the validation set to a corpus with extreme style feature would largely change their semantics.

\begin{table*}[ht]
\centering
\captionsetup{justification=centering}
\caption{Sampled out-of-domain sentences transferred by NS algorithm.}
\label{tab:ood_visualization}
 \begin{adjustbox}{max width=\textwidth}
\begin{tabular}{ll|ll}

\hline

\multicolumn{2}{c}{Yelp (Tense)} & \multicolumn{2}{c}{Amazon (Attitude)}\cr
\hline
\hline
Source&she did not finish the liver . & Source&the edger function did not work well for me .
\cr
Past&she did not finish the liver . & Neg&the edger function did not work well for me .
\cr
Pres&she does not finish the liver . & Pos&the edger function work well for me .
\cr
\hline
Source&however , i think i would try somewhere else to dine . & Source&my daughter was just frustrated with this toy .\cr
Past&however , i thought i would try somewhere to dine . & Neg&my daughter was just upset with this toy .\cr
Pres&however , i think i would try somewhere else to dine . & Pos&my daughter was just happier with this toy .\cr
\hline
\end{tabular}
\end{adjustbox}
\end{table*}

Since the trade-off between the transfer ability and content preservation can be controlled, it is hard to select one balanced point to fully characterize the performance of our method. As a complement, we suggest to use Mean as an overall performance measure, which is more stable than G-Score as observed in Fig. \ref{fig:line}. Table \ref{tab:performance} shows the performance of our methods with different recurrent architectures and baselines. As we can see, our method achieved comparable performance with two baselines while averagely outperformed them on Amazon, the benchmark with a larger vocabulary size. For an illustrative comparison, we further provide some generated samples from each method in Appendix A.

\subsection{Out-of-Domain Style Transfer} \label{sec:ood}
\begin{table}[ht]
\centering
\caption{Performance of NS algorithm on out-of-domain style transfer tasks.}\label{tab:ood}
 \begin{adjustbox}{width=0.35\textwidth}
\begin{tabular}{cccc}
\toprule
\multicolumn{4}{c}{Yelp (Tense)}\cr
\cmidrule(lr){1-4}
Acc. (Test)&BLEU&G-Score&Mean\cr
86.35(94.80)&32.64&53.09&59.49\cr
\midrule
\multicolumn{4}{c}{Amazon (Attitude)}\cr
\cmidrule(lr){1-4}
Acc. (Test)&BLEU&G-Score&Mean\cr
87.94(97.48)&22.05&44.03&55.00\cr
\bottomrule
\end{tabular}
 \end{adjustbox}

\end{table}

In the final part, we propose \textit{out-of-domain style transfer} as a much challenging task for text style transfer, where, given a corpus with style labels, the style transfer models are required to control the style of unlabeled sentences coming from out-of-domain corpora. For validation of NS algorithm's performance on this task, we use the Yelp with attitude labels only and Amazon with tense labels only to control their style on the other pair of attributes which is not observed by them. In other words, we would transfer tenses of sentences in Yelp with the $(N, S)$ operators prepared from Amazon and vice versa. 

In this scenario, we only need a slight modification on our proposed method in Sec. \ref{sec:transfer}, that is, to train the seq2seq model on Yelp $\cup$ Amazon with two style classifiers, namely attitude classifier on Yelp and tense classifier on Amazon. After preparing the style transform operators on the other domain, it is straightforward to out-of-domain style transfer. The results are reported in Table \ref{tab:ood}. It is worth to notice, even though the validation set is unlabeled in the style attribute we want to transfer, our NS algorithm can still achieve superior performance in both cases, which further validated the flexibility of style matrix perspective and the effectiveness of NS algorithm. We also provide some illustrative results in Table \ref{tab:ood_visualization}. Noticeably, the capability of out-of-domain style transfer allows us to leverage several corpus annotated with single style attributes for controlling multiple styles on each corpus.

\section{Related work}
\noindent\textbf{Unparalled Text Style Transfer.} A major proportion of works proposed to learn the style-independent semantic representations of sentences for downstream transfer tasks \citep{fu2018style,shen2017style,hu2017toward,chen2018adversarial}. These works minimized reconstruction loss of a variational autoencoder \citep{kingma2013auto} to compress the sentences and align the distributions of these vectors by adversarial training. Some other works utilized heuristic transformation to accomplish style transfer by explicitly dividing the sentence into semantic words and style words \citep{li2018delete,xu2018unpaired}. Essentially different from these previous works, our work focuses on studying how seq2seq models perceive language styles and the competitive performance of our proposed style transfer algorithm is therefore better to be considered as an implicit justification to our style matrix view on language style.

\noindent\textbf{Style in Other Domains.} Style phenomenon is also studied in other domains, especially in computer vision. The groundbreaking works by \citet{gatys2015texture,gatys2016image} showed the Gram matrices of the feature maps extracted by a pre-trained convolution neural network are able to capture the visual style of an image, which was immediately followed by numerous works have been developed to transfer the style by matching the generated Gram matrices (e.g. \citealt{ulyanov2016texture,ulyanov2017improved, johnson2016perceptual,chen2017stylebank}) and \citet{li2017demystifying} theoretically proves that it's equivalent to minimize the maximum mean discrepancy of two distributions.

\section{Conclusion}

In this paper, we have investigated the style matrix encoded by seq2seq models as an informative delegate to language style. The notion of style matrix conforms well to human experiences and existing linguistic theories on language style. In practice, we have also proposed NS algorithm as a plug-and-play solution to unpaired text style transfer which achieved competitive transfer quality with the state-of-the-art methods and meanwhile showed superior flexibility in various use cases. In the future, we plan to discuss how the quality of semantic vectors impacts the informativeness of style matrix and study what is encoded in higher-order statistics of semantic vectors.

\clearpage

\bibliography{emnlp-ijcnlp-2019}
\bibliographystyle{acl_natbib}
\appendix
\clearpage
\onecolumn
\section{Omitted Experimental Details}
\subsection{Dataset Statistics}
We provide the statistics of Amazon and Yelp datasets we used in experiments in the following table.
\begin{table}[ht]
\centering
\captionsetup{justification=centering}
\caption{Datasets statistics.} \label{tab:statistics}
\scalebox{0.75}{
\begin{tabular}{c|c|c|c|c|c}
\hline
Dataset&\makecell[c]{Vocabulary\\Size}&Attributes&Train&Dev&Test\\
\hline
\multirow{4}*{Yelp}&\multirow{4}*{9603}&Positive&173K&37614&76392\\
\cline{3-6}
~&~&Negative&263K&24849&50278\\
\cline{3-6}
~&~&Past&82K&10682&22499\\
\cline{3-6}
~&~&Present&117K&13455&26362\\
\hline
\multirow{4}*{Amazon}&\multirow{4}*{33640}&Positive&100K&38319&957\\
\cline{3-6}
~&~&Negative&100K&35899&916\\
\cline{3-6}
~&~&Past&63K&5K&1K\\
\cline{3-6}
~&~&Present&129K&5K&1K\\
\hline

\end{tabular}}

\end{table}

\subsection{Procedures to Produce Fig. 1}
We calculated four style matrices to the corpora associated with different ratings with the style matrix extraction method introduced in Sec.4.3.1 to visualize their similarities and differences of styles the style matrix captured. We picked their first and second eigenvectors corresponding to the first two eigenvalues and applied dimension reduction to them with Multi-Dimensional Scaling (MDS) to the 2-D plane.

\subsection{Sampled Sentences with Different Style Transfer Methods}

\begin{longtable}{ll}
\toprule
\multicolumn{2}{c}{From negative to postitive (Yelp)}\cr
\cmidrule(lr){1-2}
Source&the food tasted awful .\cr
Cross Aligned&the food is amazing .\cr
Style Embedding&the food tasted awful .\cr
Ours&the food tasted amazing .\cr
\cr
Source&i love the food ... however service here is horrible .\cr
Cross Aligned&i love the food here is great service great .\cr
Style Embedding&i love the food ... however service here is horrible .\cr
Ours&i love the food , service here is great .\cr
\cr
Source&customer service is horrible , and their prices are well above internet pricing .\cr
Cross Aligned&great service , and prices are great , well quality their people .\cr
Style Embedding&customer service is horrible , and their prices are well above internet pricing .\cr
Ours&customer service is excellent and their prices are nice internet pricing .\cr

\midrule
\multicolumn{2}{c}{From positive to negative (Yelp)}\cr
\cmidrule(lr){1-2}
Source&lol , we all love love love this deli .\cr
Cross Aligned&then , we love , but i love this salon .\cr
Style Embedding&lol , we all love love love this deli .\cr
Ours&lol , everyone really do n't love this deli .\cr
\cr
Source&one of the best service experiences i 've ever had .\cr
Cross Aligned&one of the time i would i had ever had to .
\cr
Style Embedding&one of the best service experiences i 've ever had .
\cr
Ours&one of the worst service experiences i 've ever had .\cr
\cr
Source&would definitely recommend this place for anyone looking for a good sandwich .\cr
Cross Aligned&would not recommend this place for a good for \_num\_ for a food .\cr
Style Embedding&would definitely recommend this place for anyone looking for a good sandwich .\cr
Ours&would not recommend this place for anyone looking for a good sandwich .\cr

\midrule
\multicolumn{2}{c}{From negative to postitive (Amazon)}\cr
\cmidrule(lr){1-2}
Source&there are so many expensive natural products out there .\cr
Cross Aligned&there are more than other than there are there .\cr
Style Embedding&there are so many expensive natural products out there .\cr
Ours&there are so many natural products out there .\cr
\cr
Source&toaster looks better but performs far worse than \$ ones i ve had .\cr
Cross Aligned&the price works great as far as far i have ever ordered .
\cr
Style Embedding&toaster looks better but performs much similar awesome if i were this phone .
\cr
Ours&toaster looks better but far better than \$ ones i ve had .\cr

\midrule
\multicolumn{2}{c}{From positive to negative (Amazon)}\cr
\cmidrule(lr){1-2}
Source&the product was delivered on the agreed date .\cr
Cross Aligned&the product was was on the same problem .\cr
Style Embedding&the product was delivered on the page said .\cr
Ours&the product was delivered on the expiration date .\cr
\cr
Source&i bought a second one with the same wonderful results .
\cr
Cross Aligned&i bought a replacement one of the same $<$unk$>$ problem .\cr
Style Embedding&i bought a second one with the same wonderful results .\cr
Ours&i bought a second one with the same results .\cr

\bottomrule
\end{longtable}
\label{tabX}

\subsection{More Samples by NS Algorithm on Out-of-Domain Style Transfer}

\begin{table*}[h]
\centering

\begin{threeparttable}
\begin{tabular}{ll}
\toprule

\multicolumn{2}{c}{Yelp on Tense}\cr
\cmidrule(lr){1-2}

Source&staff are nice and friendly .\cr
Past&staff were nice and friendly .\cr
Pres&staff are nice and friendly .\cr
\cr
Source&i never realized the beauty of the desert until i moved here !\cr
Past&i never realized the beauty of the desert until i moved here !\cr
Pres&i never assume the beauty of the desert i 'm coming here !\cr

\midrule

\multicolumn{2}{c}{Amazon on Attitude}\cr
\cmidrule(lr){1-2}

Source&yep , i thought \$ was a pretty good price .\cr
Neg&yep , i thought \$ was not a pretty good price .\cr
Pos&yep , i thought \$ was pretty good .\cr
\cr
Source&i even tried it once and it is absolutely delicious .\cr
Neg&i even tried it once and it is absolutely just seasoned .\cr
Pos&i even tried it once and it is delicious !\cr
\bottomrule

\end{tabular}
\end{threeparttable}
\end{table*}

\end{document}